\documentclass[twoside,11pt]{article}

\usepackage[abbrvbib, preprint, nohyperref]{jmlr2e}

\usepackage[utf8]{inputenc}
\usepackage[T1]{fontenc}
\usepackage{booktabs}
\usepackage{amsfonts}
\usepackage{nicefrac}
\usepackage{microtype}
\usepackage{xcolor}
\usepackage{times}
\usepackage{graphicx}      
\usepackage{xspace}
\usepackage{multirow}
\usepackage{mymacros}
\usepackage{enumitem}
\usepackage{colortbl}
\usepackage{enumitem}
\pgfplotsset{compat=1.18}
\usepackage{wrapfig}
\usepackage{url}
\usepackage[hidelinks]{hyperref}
\hypersetup{breaklinks=true}

\newcommand{\igpucb}{\texttt{IGP-UCB}\@\xspace}

\newcommand{\klucb}{\texttt{KL-UCB}\@\xspace}

\usepackage{lastpage}
\jmlrheading{23}{2024}{1-\pageref{LastPage}}{1/21; Revised 5/22}{9/22}{21-0000}{Marco Mussi, Simone Drago and Alberto Maria Metelli}
\ShortHeadings{Open Problem: Tight Bounds for Kernelized MABs with Bernoulli Rewards}{Mussi, Drago and Metelli}
\firstpageno{1}

\begin{document}

\setlength{\abovedisplayskip}{5pt}
\setlength{\belowdisplayskip}{5pt}

\title{Open Problem: Tight Bounds for \\ Kernelized Multi-Armed Bandits with Bernoulli Rewards}
\author{\name Marco Mussi \email marco.mussi@polimi.it \\
       \name Simone Drago \email simone.drago@polimi.it \\
       \name Alberto Maria Metelli \email albertomaria.metelli@polimi.it \\
       \addr Politecnico di Milano, Piazza Leonardo da Vinci 32, Milan, Italy}
\editor{Preprint}

\maketitle

\begin{abstract}
We consider Kernelized Bandits (KBs) to optimize a function $f : \mathcal{X} \rightarrow [0,1]$ belonging to the Reproducing Kernel Hilbert Space (RKHS) $\mathcal{H}_k$. Mainstream works on kernelized bandits focus on a subgaussian noise model in which observations of the form $f(\xs_t)+\epsilon_t$, being $\epsilon_t$ a subgaussian noise, are available~\citep{chowdhury2017kernelized}. Differently, we focus on the case in which we observe realizations $y_t \sim \text{Ber}(f(\xs_t))$ sampled from a Bernoulli distribution with parameter $f(\xs_t)$. While the Bernoulli model has been investigated successfully in multi-armed bandits~\citep{garivier2011kl}, logistic bandits~\citep{FauryAJC22}, bandits in metric spaces~\citep{magureanu2014lipschitz}, it remains an open question whether tight results can be obtained for KBs. This paper aims to draw the attention of the online learning community to this open problem.
\end{abstract}


\section{Introduction}
\label{sec:intro}

In this work, we present three open problems related to Kernelized Bandits (KBs, \citealt{chowdhury2017kernelized}) for optimizing a function $f : \mathcal{X} \rightarrow [0,1]$ belonging in the Reproducing Kernel Hilbert Space (RKHS) $\mathcal{H}_k$. We assume to observe samples $y_t \sim \text{Ber}(f(\xs_t))$ from a Bernoulli distribution with parameter $f(\xs_t)$. In the following, we revise the literature about Bernoulli observations coupled with different bandit structures and the subgaussian noise model for KBs.

\vspace{.1cm}

\noindent\textbf{Bernoulli Samples.}~~~\citet{garivier2011kl} developed the first optimal algorithm (\klucb) for regret minimization in Multi-Armed Bandits (MABs) with Bernoulli rewards (and no structure on the arms). \klucb leverages \emph{optimism} and a concentration bound based on the Kullback-Leibler divergence~\citep[KL,][]{kullback1951information} succeeding in asymptotically matching the lower bound~\citep{lai1985asymptotically}. Several works extended MABs with Bernoulli rewards to account for structure, including metric spaces~\citep{magureanu2014lipschitz} and linear (logistic) models~\citep{FauryAJC22}.

\vspace{.1cm}

\noindent\textbf{Kernelized Bandits.}~~~The seminal work \citep{srinivas2010gaussian} introduce \gpucb, the first no-regret solution based on Gaussian  Processes~\citep[GPs,][]{rasmussen2006gaussian}. \gpucb enjoys regret guarantees both in the cases when $f$ is indeed sampled from a GP and when $f$ belongs to a suitable RKHS (\emph{agnostic} case).
\citep{chowdhury2017kernelized} derive an improved version of \gpucb, called \igpucb, working under subgaussian noise model. The analysis is based on a novel \emph{self-normalized concentration inequality} for subgaussian samples $f(\xs_t)+\epsilon_t$ that extends and generalizes that of~\citep{abbasi2011improved} for linear models. These solutions can be adapted to learn also in the presence of Bernoulli rewards\footnote{A Bernoulli random variable is $\lambda$-subgaussian, with $\lambda= 1/2$.} at the price of (possibly) looser guarantees.

These results are possible thanks to specifically designed concentration bounds, which are (almost) optimal for their specific settings. As we can notice from Table~\ref{tab:summary}, the only missing solution is the one to learn with kernelized structure in the presence of Bernoulli rewards. The goal of this work is to raise the attention of the online learning community on this gap in the literature.

\begin{table}[t!]
    \centering
    \renewcommand{\arraystretch}{1.35}
    \resizebox{\linewidth}{!}{
    \small
    \begin{tabular}{|c||c|c|}
    \hline 
    & \emph{Subgaussian} & \cellcolor{vibrantCyan!15} \emph{Bernoulli} \\
    \hline
    \hline
    \emph{No Structure} & \citealt{lattimore2020bandit} (Corollary~5.5) & \cellcolor{vibrantCyan!15} \citealt{garivier2011kl} (Theorem~10) \\
    \hline
    \emph{Linear} & \citealt{abbasi2011improved} (Theorem~2) & \cellcolor{vibrantCyan!15} \citealt{FauryAJC22} (Proposition~3) \\
    \hline
    \emph{Metric Space} & \citealt{kleinberg2008multi} (Theorem~4.2) & \cellcolor{vibrantCyan!15} \citealt{magureanu2014lipschitz} (Theorem~2) \\
    \hline
    \cellcolor{vibrantRed!15} \emph{RKHS} & \cellcolor{vibrantRed!15} \citealt{chowdhury2017kernelized} (Theorem~2) & \cellcolor{red!50!blue!15} \textbf{{Open Problem}} \\
    \hline
    \end{tabular}}
    \caption{Summary of the state-of-the-art in concentration bounds.}
    \label{tab:summary}
\end{table}

\section{Problem Formulation}
\label{sec:problemformulation}

In this section, we describe the setting, the learning problem, and the considered assumptions.

\vspace{.1cm}

\noindent\textbf{Setting.}~~~We consider the problem of sequentially maximizing a fixed and unknown function $f:\mathcal{X} \rightarrow [0,1]$ over a decision set $\mathcal{X} \subseteq \mathbb{R}^d$ (also called action set). At every round $t \in \dsb{T} \coloneqq \{1,\dots, T\}$, being $T \in \mathbb{N}$ the learning horizon, the algorithm $\mathfrak{A}$ chooses an action $\xs_t \in \mathcal{X}$ based on the history of past observations $\mathcal{G}_t \coloneqq \{(\xs_s,y_s)\}_{s\in\dsb{t-1}}$ and observes a random variable $y_t \sim \text{Ber}(f(\xs_t))$, where $\text{Ber}(p)$ denotes a Bernoulli distribution with parameter $p \in [0,1]$.

\vspace{.1cm}

\noindent\textbf{Learning Problem.}~~~The goal of the learning algorithm $\mathfrak{A}$ is to minimize the \emph{regret}:
\begin{equation}
    R_T (\mathfrak{A}) \coloneqq T \, f(\xs^\star) -   \sum_{t \in\dsb{T}} f(\xs_t) \qquad \text{where} \qquad \xs^\star \in \argmax_{\xs \in \mathcal{X}} f(\xs).
\end{equation}

\noindent\textbf{Regularity Conditions.}~~~We consider the frequentist-type regularity assumption that is usually employed in KBs~\citep{srinivas2010gaussian,chowdhury2017kernelized}. Let $\mathcal{H}_k$ be a RKHS induced by the kernel function $k:\mathcal{X \times X} \rightarrow \mathbb{R}$ so that every function $h \in \mathcal{H}_k$ satisfies the \emph{reproducing property} $h(x) = \langle h, k(\cdot,x) \rangle_{\mathcal{H}_k}$, where $\langle \cdot,\cdot \rangle_{\mathcal{H}_k}$ is the inner product defined on the space $\mathcal{H}_k$. We denote with $\|h\|_{\mathcal{H}_k} = \sqrt{\langle h,h \rangle_{\mathcal{H}_k}}$ the RKHS norm. We enforce the following standard assumption prescribing that $f $ belongs to the RKHS with bounded norm.

\begin{ass}[Regularity Conditions]\label{ass} $f $ belongs to the RKHS, i.e., $f \in \mathcal{H}_k$, and: 
\begin{itemize}[noitemsep, topsep=0pt]
    \item the function $f$ has a bounded RKHS norm, i.e., $\| f\|_{\mathcal{H}_k} \le B < +\infty$ ($B$ is known);
    \item the kernel function $k$ is bounded, i.e., $k(\xs,\xs) \le 1$ for every $\xs \in \mathcal{X}$.
\end{itemize}
\end{ass}

\section{Open Problems}

\subsection{Open Problem 1: Estimation}\label{sec:op1}
\begin{center}
\emph{Can we effectively estimate $f(\xs)$ in a new point $\xs \in \mathcal{X}$ based on the \\ history  of past observations $\mathcal{G}_t \coloneqq \{(\xs_s,y_s)\}_{s\in\dsb{t-1}}$ where $y_s$ are Bernoulli samples?}
\end{center}
When the observations are of the form  $y_t = f(\xs_t) + \epsilon_t$ with $\epsilon_t$ being a $\lambda$-subgaussian noise, the standard approach consists in resorting to GPs. We consider a prior GP model $\mathcal{GP}(0,k(\cdot,\cdot))$ for function $f$ and a Gaussian likelihood model $\mathcal{N}(0,\nu^2)$ for the noise $\epsilon_t$.\footnote{The GP model is used for estimation and the true $f$ may not be sampled from the GP~\citep{chowdhury2017kernelized}.} Given the history $\mathcal{G}_t \coloneqq \{(\xs_s,y_s)\}_{s\in\dsb{t-1}}$, the posterior distribution of $f$ is $\mathcal{GP}(\mu_t(\cdot),k_t(\cdot,\cdot))$, where, for every $\xs,\xs'\in\mathcal{X}$:
\begin{align*}
    \mu_t(\xs) \coloneqq \mathbf{k}_t(\xs)^\top (\mathbf{K}_t + \nu^2 \mathbf{I})^{-1} \mathbf{y}_t, \quad \ \
    k_t(\xs,\xs') \coloneqq k(\xs,\xs) - \mathbf{k}_t(\xs)^\top (\mathbf{K}_t + \nu^2 \mathbf{I})^{-1} \mathbf{k}_t(\xs'),
\end{align*}
where $ \mathbf{k}_t(\xs) \coloneqq (k(\xs_1,\xs),\dots,k(\xs_{t},\xs))^\top$, $\mathbf{K}_t \coloneqq (k(\xs_i,\xs_j))_{i,j \in \dsb{t}}$, and $\mathbf{y}_t \coloneqq (y_1,\dots,y_t)^\top$. This allows to have the estimate $\mu_t(\xs)$ and an index of uncertainty $\sigma^2_t(\xs) \coloneqq k_t(\xs,\xs)$ of the estimate.

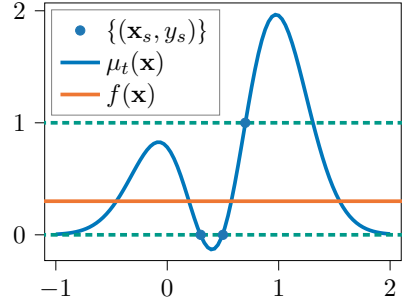
\begin{wrapfigure}{r}{0.36\textwidth}
    \centering
    \vspace{-.2cm}
    \resizebox{0.36\textwidth}{!}{
\begin{tikzpicture}

\definecolor{darkgray176}{RGB}{176,176,176}
\definecolor{green}{RGB}{0,128,0}
\definecolor{orange}{RGB}{255,165,0}
\definecolor{steelblue31119180}{RGB}{31,119,180}

\begin{axis}[
height=5.5cm,
width=7cm,
legend cell align={left},
legend style={fill opacity=0.8, draw opacity=1, text opacity=1, draw=gray, at={(.492,0.975)}},
tick align=outside,
tick pos=left,
x grid style={darkgray176},
xmin=-1.1, xmax=2.1,
xtick style={color=black},
y grid style={darkgray176},
ymin=-0.23, ymax=2.07,
ytick style={color=black}
]
\addplot [draw=steelblue31119180, fill=steelblue31119180, mark=*, only marks]
table{%
x  y
0.3 0
0.5 0
0.7 1
};
\addlegendentry{$\{(\xs_s,y_s)\}$}
\addplot [very thick, vibrantBlue, ultra thick]
table {%
-1 0.0049757381553741
-0.96969696969697 0.00671202691792301
-0.939393939393939 0.00898176069334602
-0.909090909090909 0.0119223508425118
-0.878787878787879 0.0156976443318478
-0.848484848484849 0.0205001561186622
-0.818181818181818 0.0265525416515817
-0.787878787878788 0.0341079379233117
-0.757575757575758 0.0434487493245212
-0.727272727272727 0.0548834181528066
-0.696969696969697 0.0687407079765061
-0.666666666666667 0.0853610508440759
-0.636363636363636 0.105084576268139
-0.606060606060606 0.128235559676832
-0.575757575757576 0.155103207014667
-0.545454545454545 0.185918933107855
-0.515151515151515 0.220830591874683
-0.484848484848485 0.259874467603482
-0.454545454545455 0.302946222037458
-0.424242424242424 0.349772387685461
-0.393939393939394 0.399884371776429
-0.363636363636364 0.452597249310689
-0.333333333333333 0.506995835234777
-0.303030303030303 0.561930591417172
-0.272727272727273 0.616025803580824
-0.242424242424242 0.667702124446859
-0.212121212121212 0.715215002981001
-0.181818181818182 0.756709704665671
-0.151515151515151 0.790292594861633
-0.121212121212121 0.814117151546095
-0.0909090909090908 0.826481864538216
-0.0606060606060606 0.825935857547281
-0.0303030303030303 0.811386846040649
0 0.782205036555999
0.0303030303030303 0.738315899646625
0.0606060606060606 0.680274514970847
0.0909090909090908 0.609314474696294
0.121212121212121 0.527365186411534
0.151515151515152 0.437032840414735
0.181818181818182 0.341542249404869
0.212121212121212 0.244639130342259
0.242424242424242 0.15045502987881
0.272727272727273 0.0633398090010515
0.303030303030303 -0.0123308124402914
0.333333333333333 -0.0723629133614043
0.363636363636364 -0.112955809012409
0.393939393939394 -0.130904886763525
0.424242424242424 -0.12378334372144
0.454545454545455 -0.0900911140202121
0.484848484848485 -0.0293609364575245
0.515151515151515 0.0577860416203784
0.545454545454545 0.169639666748095
0.575757575757576 0.303456275798168
0.606060606060606 0.455567786912954
0.636363636363636 0.621538441439177
0.666666666666667 0.796360218647585
0.696969696969697 0.974675551339448
0.727272727272727 1.15101439506975
0.757575757575758 1.32003205394061
0.787878787878788 1.47673447067499
0.818181818181818 1.61667889781337
0.848484848484849 1.73613985793698
0.878787878787879 1.83223289248691
0.909090909090909 1.90299156926175
0.939393939393939 1.94739632722236
0.96969696969697 1.96535674596525
1 1.95765152171119
1.03030303030303 1.9258326377805
1.06060606060606 1.87210181346442
1.09090909090909 1.79916823760168
1.12121212121212 1.71009683745755
1.15151515151515 1.60815594876541
1.18181818181818 1.49667233226533
1.21212121212121 1.37890015060982
1.24242424242424 1.2579089195297
1.27272727272727 1.13649372467776
1.3030303030303 1.01710928757142
1.33333333333333 0.901827888282775
1.36363636363636 0.792319800357804
1.39393939393939 0.689853826116305
1.42424242424242 0.595314768495071
1.45454545454545 0.509234240949107
1.48484848484848 0.431831077491555
1.51515151515152 0.363057720252905
1.54545454545455 0.302649279545123
1.57575757575758 0.250172423275505
1.60606060606061 0.20507180067118
1.63636363636364 0.16671228632553
1.66666666666667 0.134415899251121
1.6969696969697 0.107492772617608
1.72727272727273 0.0852659986142148
1.75757575757576 0.0670905352768391
1.78787878787879 0.0523666333187811
1.81818181818182 0.0405484237776158
1.84848484848485 0.0311484100947476
1.87878787878788 0.0237386432466925
1.90909090909091 0.017949339812343
1.93939393939394 0.013465644810429
1.96969696969697 0.0100231573649367
2 0.00740273968413156
};
\addlegendentry{$\mu_t(\xs)$}
\addplot [ultra thick, vibrantOrange]
table {%
-1.15 0.3
2.15 0.3
};
\addlegendentry{$f(\xs)$}
\addplot [ultra thick, vibrantTeal, densely dashed]
table {%
-1.15 0
2.15 0
};
\addplot [ultra thick, vibrantTeal, densely dashed]
table {%
-1.15 1
2.15 1
};
\end{axis}

\end{tikzpicture}}%
    \vspace{-.5cm}
    \caption{GP estimate example of $f \in [0,1]$.}
    \label{fig}
    \vspace{-.6cm}
\end{wrapfigure}
This approach can be employed when $y_t \sim \text{Ber}(f(\xs_t))$, since a Bernoulli variable is $1/2$-subgaussian. However, the drawback is that  $\mu_t(\xs)$ is not guaranteed to lie in $[0,1]$, although the true $f(\xs) \in [0,1]$, being the parameter of a Bernoulli distribution (Figure~\ref{fig}). 

A first attempt to fix this consists of keeping the prior $\mathcal{GP}(0,k(\cdot,\cdot))$ for the unknown function $f$ and \emph{change the likelihood model} to a Bernoulli one. However, this attempt is unsuccessful since the posterior computation would require evaluating the conditional distribution $\Pr({y}_t|{f}(\xs_t))$ that is not well defined when $f \sim \mathcal{GP}(0,k(\cdot,\cdot))$ since $f(\xs_t)$ may take values outside $[0,1]$.

A second attempt consists in \emph{changing both the prior and the likelihood model} to overcome the \quotes{incompatibility} between the GP and the Bernoulli likelihood model. Aiming for a conjugate prior update, we should use a \emph{Beta prior and a Bernoulli likelihood} model. However, as noted in~\citep{rolland2019efficient}, enforcing correlation with Beta distributions is challenging differently from the Gaussian case. The notion of \quotes{Beta process} was introduced in survival analysis but displays a too-limited modeling power~\citep{hjort1990nonparametric,paisley2009nonparametric}. Furthermore, there is no consensus on one definition of \emph{multivariate Beta} distribution. A common approach \citep{westphal2019simultaneous} bases on a Dirichlet distribution $\text{Dir}(\bm{\zeta})$ defined over the support $\{0,1\}^{2^t}$  (with parameter $\bm{\zeta} =(\zeta_1,\dots,\zeta_{2^t})^\top \in \mathbb{R}_{\ge 0}^{2^t}$) from which to sample  a probability vector $\mathbf{p}_t = (p_1,\dots,p_{2^t})^\top \sim\text{Dir}(\bm{\zeta})$ needed to define the multivariate Beta variable as $\bm{\theta} = \mathbf{H}_t \mathbf{p}_t$ where $\mathbf{H}_t = (\text{bin}(0)|\dots |\text{bin}(2^t))$ and $\text{bin}(n)$ is the binary encoding of number $n$. Although this allows for a simple posterior calculation, it is completely unstructured and does not allow embedding the structure enforced by the kernel $k$.

These attempts focus on deriving a \quotes{proper} Bayesian update. Since even for the subgaussian KBs, GPs are just an estimation tool, we may consider \emph{non-Bayesian} updates. \citep{goetschalckx2011continuous} proposes a sample-sharing method in which samples contribute weighted by the kernel $k$:
\begin{align*}
    \alpha_t(\xs) \coloneqq \alpha_0 + \sum_{s\in\dsb{t}} y_s k(\xs,\xs_s),  \qquad \beta_t(\xs) \coloneqq \beta_0 + \sum_{s\in\dsb{t}} (1-y_s) k(\xs,\xs_s). 
\end{align*}
This approach has convergence guarantees when $f$ is Lipschitz continuous.
Other approaches leverage on Logistic Gaussian Processes~\citep{leonard1978density} or Gaussian Process Copulas~\citep{wilson2010copula}, and they all involve non-Bayesian updates due to the analytical intractability.

\subsection{Open Problem 2: Concentration}\label{op2}
\begin{center}
\emph{Can we derive concentration guarantees for the deviation $|f(\xs) - \mu_t(\xs)|$ \\ (being $\mu_t(\xs)$ a suitable estimator of $f(\xs)$) which is tight for the Bernoulli observations?}
\end{center}

For the $\lambda$-subgaussian case, by resorting to the GP-based estimator presented in Section~\ref{sec:op1} \citep{srinivas2010gaussian}, it is possible, under Assumption~\ref{ass}, to achieve the following concentration bound for the deviation that holds w.p. $1-\delta$ simultaneously for every $t \ge 1$ and $\xs \in \mathcal{X}$:
\begin{align}\label{bound}
    |\mu_{t-1}(\xs) - f(\xs)| \le \left( B + \lambda \sqrt{2\left(\gamma_{t-1} + 1 + \log(1/\delta) \right) }\right) \sigma_{t-1}(\xs),
\end{align} 
where $\gamma_{t} = \max_{\mathcal{A} \subset \mathcal{X} : |\mathcal{A}| = t} I(\mathbf{y}_{\mathcal{A}};\mathbf{f}_{\mathcal{A}})$ is the \emph{maximum information gain} at time $t$. This result is obtained by deriving a \emph{self-normalized concentration inequality} that bounds a suitable weighted norm of the noise process $(\epsilon_1,\dots,\epsilon_{t-1})^\top$~\citep[][Theorem 1]{chowdhury2017kernelized}. Clearly, Equation~\eqref{bound} holds for Bernoulli distributions too, being them subgaussian with $\lambda = 1/2$. 

While Equation~\eqref{bound} is likely tight for subgaussian observations, it fails to capture the stronger concentration rate that characterizes Bernoulli random variables. Indeed, in the unstructured case (i.e., MABs with no correlation between the arms), \citet{garivier2011kl} obtain the stronger concentration bound, holding w.p. $1-\delta$ for a fixed $\xs \in \mathcal{X}$ and simultaneously for every $t \ge 0$:
\begin{align}
    \begin{array}{ll}
& \text{Let:} \ \ \quad u_{t}(\xs) \coloneqq \max\{ q \ge \mu_{t}(\xs) : N_{t}(\xs) d(\mu_{t}(\xs),q) \le c_1\log (t/\delta)  + c_2\log\log (t/\delta)  \} , \\
    &  \text{then:} \, \quad f(\xs) \le  u_{t}(\xs),
    \end{array} \label{gc}
\end{align}
where $d(a,b) = a\log(a/b) + (1-a) \log((1-a)/(1-b))$ for $a,b \in [0,1]$ is the Bernoulli KL-divergence, $\mu_{t}(\xs) = \sum_{s\in\dsb{t}} y_s \mathbf{1}\{\xs_s=\xs\} / N_t(\xs)$, $N_t(\xs) = \sum_{s\in\dsb{t}} \mathbf{1}\{\xs_s=\xs\}$, and $c_1,c_2>0$ are universal constants. Equation~\eqref{gc} evaluates the distance using the KL-divergence $d(\cdot,\cdot)$ between Bernoulli parameters and, therefore, delivers a stronger concentration bound compared to that of Equation~\eqref{bound}. The derivation of this result \citep{garivier2011kl} makes use of a \emph{martingale}-based argument deeply depending on the moment-generating function of the Bernoulli distribution that seems not to be easily extensible to the KB setting in which correlation among arms is present.

\subsection{Open Problem 3: Regret Minimization}
\begin{center}
    \emph{Can we design regret minimization algorithms which achieve a $\log T$ regret guarantee, \\highlighting the dependence on $d(f(\xs),f(\xs^\star))$ when $\mathcal{X}$ is finite?}
\end{center}

Under Assumption~\ref{ass}, by applying the improved \gpucb presented in \igpucb \citep[][Theorem 2]{chowdhury2017kernelized}, we obtain a \emph{worst-case $\widetilde{\mathcal{O}}(\sqrt{T})$} regret guarantee w.p. $1-\delta$:
\begin{align*}
    R_T(\text{\igpucb}) \le \mathcal{O} \left( B \sqrt{\gamma_T} + \sqrt{T \gamma_T \left(\gamma_T + \log (1/\delta) \right)}\right).
\end{align*}
The study of \emph{instance-dependent} regret bounds for KBs is introduced in~\citep{shekhar2022instance}, focusing on the \emph{packing} properties of the RKHS and still achieving regret bounds of order $\widetilde{\mathcal{O}}(T^{\alpha})$ for some $\alpha \in (0,1)$. Here, instead, when $\mathcal{X}$ is finite, we are interested in understanding if achieving $\log T$ regret is possible for KBs with Bernoulli observations. Indeed, in the unstructured case (and $|\mathcal{X}| <+ \infty$), the \klucb~\citep{garivier2011kl} achieves the tight \emph{instance-dependent} bound:
\begin{align*}
     R_T(\text{\klucb}) \le \mathcal{O} \bigg( \sum_{\xs \in \mathcal{X}} \frac{\log T}{d(f(\xs),f(\xs^\star))} \bigg).
\end{align*}
We perceive that this should be possible since, when $|\mathcal{X}| <+ \infty$, using the trivial kernel $k(\xs,\xs') = \mathbf{1}\{\xs=\xs'\}$ for every $\xs,\xs' \in \mathcal{X}$, the KB reduces to an unstructured MAB. Furthermore, we conjecture that this possibility (at least for optimistic algorithms) is strictly related to the open problem of Section~\ref{op2}. Indeed, the bound of Equation~\eqref{gc} is specifically designed for the \klucb algorithm~\citep{garivier2011kl}.

\clearpage

\bibliography{biblio}

\end{document}